\documentclass[letterpaper, 10 pt, conference]{ieeeconf}  

\IEEEoverridecommandlockouts                              

\usepackage{graphics} 
\usepackage{epsfig} 
\usepackage{mathptmx} 
\usepackage{times} 
\usepackage{amsmath} 
\usepackage{amssymb}  
\usepackage{booktabs}
\usepackage{url}

\newcommand{\ourMethod}{HybridTM}

\title{\LARGE \bf
Hybrid Transformer-Mamba Model for 3D Semantic Segmentation
}

\author{{Xinyu Wang, Jinghua Hou, Zhe Liu, Yingying Zhu*}
\thanks{This work was supported by the NSFC (Grant 62225603).}
\thanks{This work was also supported by the NSFC (Grant U2341227). }
\thanks{Xinyu Wang and Jinghua Hou contributed
 equally to this work. *Corresponding author: Yingying Zhu.}
\thanks{Xinyu Wang is with School of Artificial Intelligence and Automation, Huazhong University of Science and Technology (HUST), Wuhan 430074, China
(e-mail: {xinyuwang@hust.edu.cn})}%
\thanks{Jinghua Hou is with School of Electronic Information and Communications, Huazhong University of Science and Technology (HUST), Wuhan 430074, China (e-mail: jhhou@hust.edu.cn)}
\thanks{Zhe Liu is with the Department of Computer Science of The University of Hong Kong (HKU), Pokfulam 999077, Hong Kong (e-mail: zheliu12@hku.hk)}
\thanks{Yingying Zhu is with School of Computer Science and Technology, Huazhong University of Science and Technology (HUST), Wuhan 430074, China (e-mail: yyzhu@hust.edu.cn)}
}

\begin{document}

\maketitle
\thispagestyle{empty}
\pagestyle{empty}

\begin{abstract}

Transformer-based methods have demonstrated remarkable capabilities in 3D semantic segmentation through their powerful attention mechanisms, but the quadratic complexity limits their modeling of long-range dependencies in large-scale point clouds. While recent Mamba-based approaches offer efficient processing with linear complexity, they struggle with feature representation when extracting 3D features. However, effectively combining these complementary strengths remains an open challenge in this field. In this paper, we propose {\ourMethod}, the first hybrid architecture that integrates Transformer and Mamba for 3D semantic segmentation. In addition, we propose the Inner Layer Hybrid Strategy, which combines attention and Mamba at a finer granularity, enabling simultaneous capture of long-range dependencies and fine-grained local features. Extensive experiments demonstrate the effectiveness and generalization of our {\ourMethod} on diverse indoor and outdoor datasets. Furthermore, our {\ourMethod} achieves state-of-the-art performance on ScanNet, ScanNet200, and nuScenes benchmarks.The code will be made available at \url{https://github.com/deepinact/HybridTM}.
\end{abstract}

\section{INTRODUCTION}

Semantic segmentation in point clouds has emerged as a pivotal task in 3D vision, enabling systems to assign semantic labels to each point within a 3D space. This capability is fundamental for applications such as autonomous driving~\cite{siam2018comparative, marcuzzi2023mask, sun2020pointmoseg, liu2022epnet++}, robotic navigation~\cite{kim2018indoor, chen2019multi}, and 3D scene understanding~\cite{behley2019semantickitti}.


Recently, transformer-based methods have demonstrated remarkable capabilities in 3D semantic segmentation through their powerful attention mechanisms. However, achieving a large receptive field in point clouds for transformers~\cite{vaswani2017attention} is impractical because of the quadratic computational complexity. To address this limitation, many works~\cite{zhao2021point, zhang2022patchformer, yang2023swin3d, wang2023octformer, wu2022point, wu2024point, park2022fast, lai2022stratified,lai2023spherical} have split point clouds into smaller groups and apply self-attention within these groups. While this strategy reduces computational cost, it inherently limits the ability to model long-range dependencies, leading to suboptimal segmentation performance. 

Building on the State Space Model (SSM), Mamba~\cite{gu2023mamba} can efficiently process large-scale point clouds with linear computational complexity. While Mamba has demonstrated impressive performance in various 2D vision tasks~\cite{zhuvision, liu2024vmamba, huang2024localmamba, pei2024efficientvmamba, ruan2024vm}, Mamba-based methods~\cite{liu2024point, zhang2024point, wang2024serialized} often struggle with feature representation when extracting 3D features, particularly in capturing fine-grained local features, which is important for precise segmentation.

\begin{figure}[t]
\centering
\includegraphics[width=0.85\linewidth]{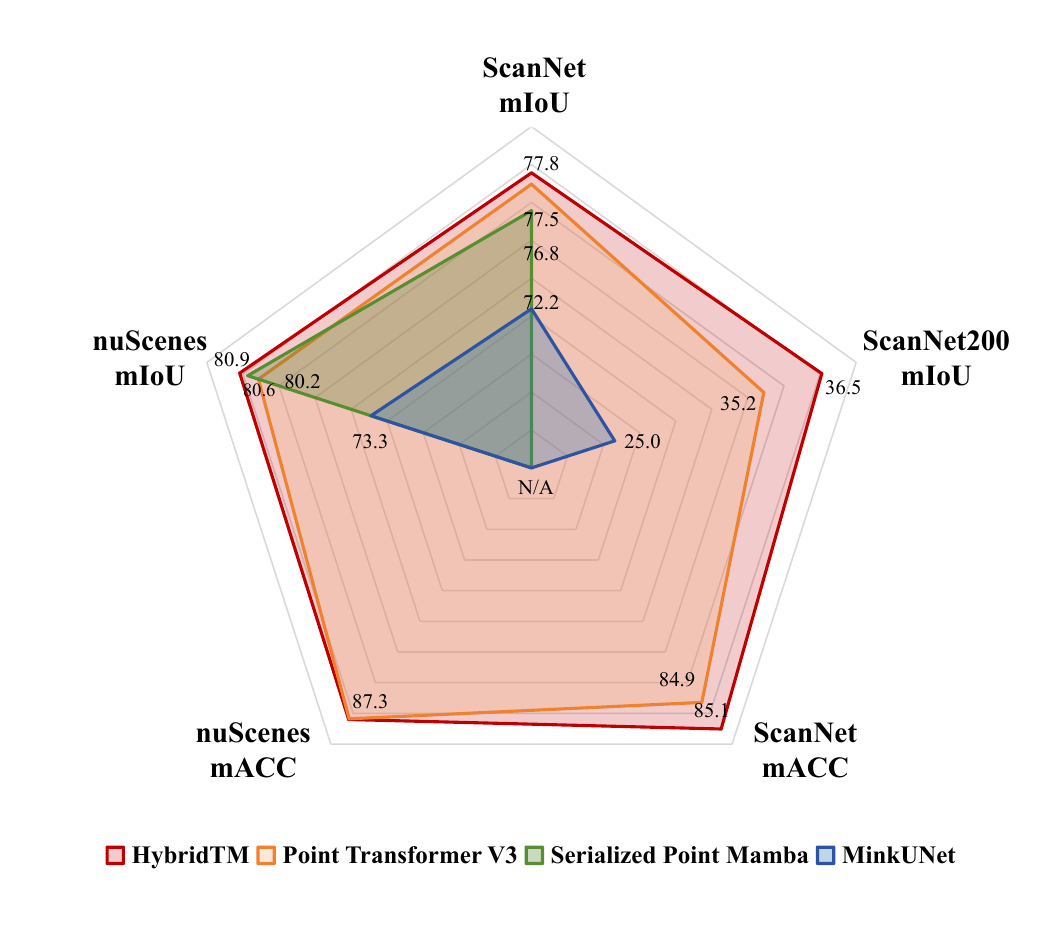} 
\vspace{-5pt}
\caption{Comparison of existing representative 3D semantic segmentation methods with different operators (e.g., Transformer~\cite{vaswani2017attention}, 3D sparse convolution~\cite{graham20183d}, and Mamba~\cite{gu2023mamba}) on ScanNet~\cite{dai2017scannet}, nuScenes~\cite{caesar2020nuscenes}, and ScanNet200~\cite{dai2017scannet} dataset. {\ourMethod} achieves superior performance on both indoor and outdoor datasets.}
\label{fig:intro}
\vspace{-10pt}
\end{figure}

While recent works have explored hybrid architectures combining attention and Mamba in NLP~\cite{lieber2024jamba} and 2D vision~\cite{hatamizadeh2024mambavision, liu2024map, chen2024maskmamba} task, there are new challenges for 3D semantic segmentation. Unlike dense 2D images, point clouds are inherently sparse and irregular. Directly applying existing hybrid strategies in 2D vision~\cite{hatamizadeh2024mambavision, liu2024map, chen2024maskmamba} to serialized sparse voxels leads to feature degradation, limiting their effectiveness in 3D semantic segmentation.

To address these challenges, we propose {\ourMethod}, the first hybrid architecture that integrates Transformer and Mamba for 3D semantic segmentation. In addition, we propose the Inner Layer Hybrid Strategy, which can simultaneously maintain the ability to capture fine-grained features and model long-range dependencies by combining these operators at a finer granularity. As shown in Fig.~\ref{fig:intro}, {\ourMethod} outperforms existing methods based on transformers, 3D sparse convolutions, and Mamba in indoor and outdoor datasets.



\begin{figure*}[t!]
\centering
\includegraphics[width=1.0\linewidth]{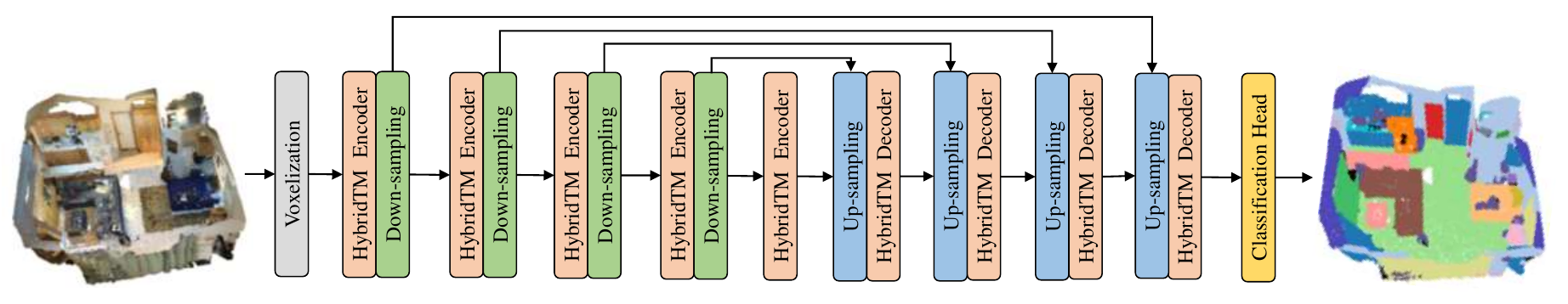}
\caption{The illustration of {\ourMethod}, which consists of {\ourMethod} encoders, {\ourMethod} decoders, down-sampling, up-sampling, and a classification head. {\ourMethod} voxelizes the input point clouds and adopts an UNet-like architecture to extract multi-scale features. Finally, extracted features are fed into a classification head for 3D semantic segmentation. }
\label{fig:overall}
\end{figure*}

In summary, our contributions are as follows. 
\begin{itemize}
    \item We propose {\ourMethod}, the first hybrid architecture that integrates Transformer and Mamba for 3D semantic segmentation. {\ourMethod} can efficiently capture the fine-grained local features and model long-range relationships with the proposed Inner Layer Hybrid Strategy. 
    \item We conducted experiments on several indoor and outdoor datasets to validate the effectiveness and generalization of {\ourMethod}. {\ourMethod} achieves state-of-the-art (SOTA) performance on Scannet~\cite{dai2017scannet}, Scannet200~\cite{dai2017scannet}, and nuScenes~\cite{caesar2020nuscenes}.
\end{itemize}

\section{RELATED WORK}

\subsection{Transformers in 3D semantic segmentation}

Transformer has been shown to be a successful architecture in many vision tasks. Therefore, many works try to adopt attention operators in 3D semantic segmentation tasks. Point Cloud Transformer (PCT)~\cite{guo2021pct} was the pioneering work, which replaced traditional self-attention with offset-attention to better capture the geometric relationships in point clouds. However, the quadratic computational complexity of Transformers makes it impractical to apply attention operators directly to entire point clouds. To address this problem, some methods~\cite{zhao2021point, zhang2022patchformer, yang2023swin3d, wang2023octformer, wu2022point, wu2024point, park2022fast, lai2022stratified, lai2023spherical, zeller2022gaussian} divide point clouds into small groups, thereby reducing the computational cost. For instance, Point Transformer~\cite{zhao2021point} refined the self-attention mechanism by vector attention, leading to a stronger feature representation. Similarly, OctFormer~\cite{wang2023octformer} utilized an octree to partition the point clouds and implemented the dilated octree attention to reduce computational cost. Despite these innovations, existing approaches often struggle to effectively model long-range dependencies inherent in 3D semantic segmentation tasks. Point Transformer v3~\cite{wu2024point} achieved better performance by enlarging the receptive field with serialized neighbor mapping, thereby improving performance. 
Nevertheless, it still fails to fully capture the extensive long-range dependencies required for optimal 3D semantic segmentation.

\subsection{Mamba in 3D vision}

Due to the linear complexity of Mamba~\cite{gu2023mamba, daotransformers}, there has been a growing interest in adapting this architecture to replace transformers across various vision tasks~\cite{zhuvision, liu2024vmamba, huang2024localmamba, pei2024efficientvmamba, ruan2024vm, hwang2024hydra, chen2024mim, wang2024omega}. Initial explorations in 2D vision, notably Vision Mamba~\cite{zhuvision} and VMamba~\cite{liu2024vmamba}, demonstrated promising results by effectively serializing 2D images into sequences for Mamba processing.

Following these successes, researchers began investigating Mamba's potential in 3D vision tasks~\cite{zhang2024point, liu2024lion, wang2024serialized, liu2024point}. Point Cloud Mamba~\cite{zhang2024point} pioneered the application of Mamba to process the point clouds, showing its capability in capturing complex geometric relationships. LION-Mamba~\cite{liu2024lion} further extended this approach to LiDAR-based 3D object detection, achieving state-of-the-art performance by modeling the long-range dependency. In the domain of 3D semantic segmentation, Serialized Point Mamba~\cite{wang2024serialized} introduced a linear complexity approach that attempts to balance local feature extraction with global context modeling.

However, despite these advances, purely Mamba-based approaches have struggled to consistently achieve state-of-the-art performance across different vision tasks. This limitation primarily stems from Mamba's inherent weakness in feature representation compared to transformer-based architectures, suggesting the need for more sophisticated architectural designs.

\subsection{Hybrid Structures for Transformer and Mamba} 



Recent efforts~\cite{lieber2024jamba, hatamizadeh2024mambavision, liu2024map, chen2024maskmamba} have begun to explore hybrid structures that combine Transformer and Mamba to leverage the strengths of each component and improve overall performance.
In NLP tasks, Jamba~\cite{lieber2024jamba} interleaved the Transformer and Mamba layers, achieving lower computational cost and better performance. 
In 2D vision tasks, recent approaches~\cite{hatamizadeh2024mambavision, liu2024map, chen2024maskmamba} combined Transformer layers with Mamba layers by the outer hybrid strategy. This strategy effectively enhances feature representations by capturing both short-range and long-range dependencies for various applications, including general computer vision~\cite{hatamizadeh2024mambavision}, backbone pretraining~\cite{liu2024map}, and image generation~\cite{chen2024maskmamba}. However, extending this hybrid strategy to 3D semantic segmentation tasks poses significant challenges, leading to suboptimal performance.
Therefore, in this paper, we propose {\ourMethod}, the first hybrid architecture that effectively combines Transformer and Mamba by our proposed Inner Layer Hybrid Strategy, achieving superior performance in 3D semantic segmentation.


\section{METHOD}

\begin{figure*}[t!]
\includegraphics[width=1.0\linewidth]{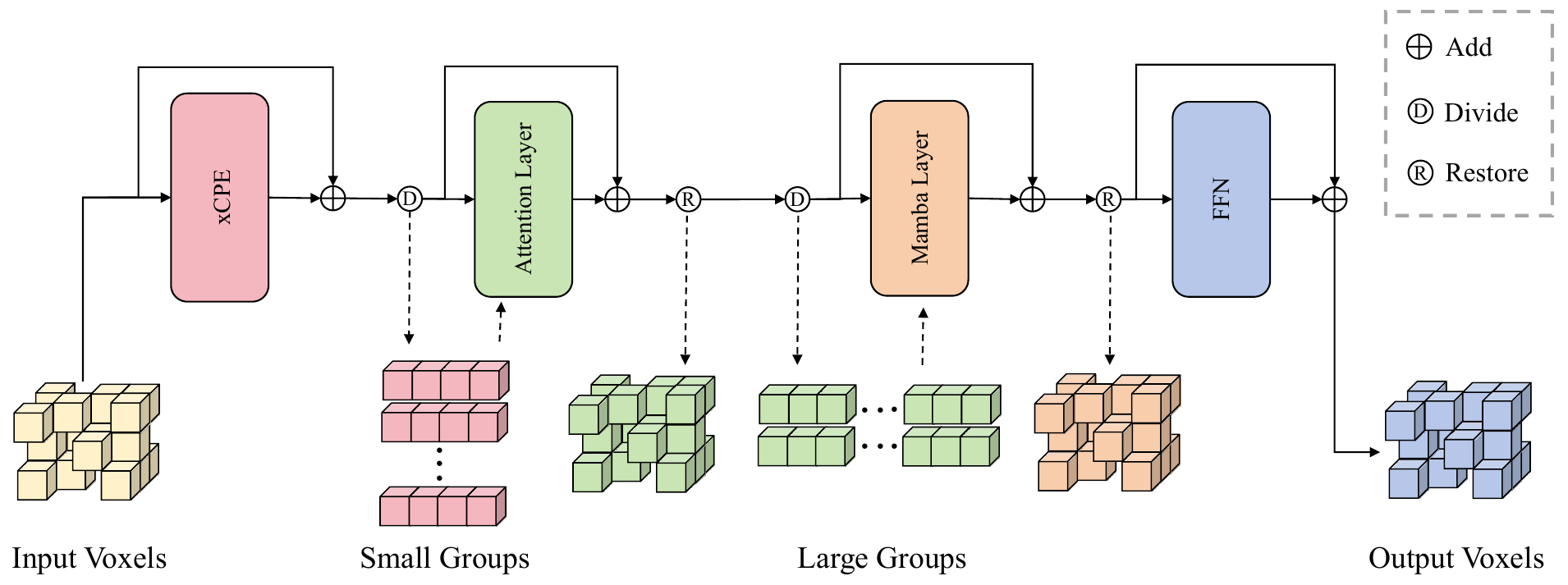}
\caption{The illustration of Hybrid Layer. The Hybrid Layer contains an xCPE, an attention layer, a Mamba layer, and an FFN layer. First, we divide the voxels enhanced by the xCPE into several small groups for the attention layer to extract fine-grained features. Then, we restore the output voxels of the attention layer to the origin shape. Then, we divide the restored voxels into several large groups for the Mamba layer to extract global features. Finally, we adopt the FFN layer to enhance the fused features for final output voxels.}
\label{fig:layer}
\end{figure*}

In this work, we propose {\ourMethod}, the first hybrid architecture for 3D semantic segmentation that synergistically integrates Transformer~\cite{vaswani2017attention} and Mamba~\cite{gu2023mamba} layers through our proposed Inner Layer Hybrid Strategy. Our architecture can simultaneously extract fine-grained local features and model long-range dependencies, enabling more accurate 3D semantic segmentation. As illustrated in Fig.~\ref{fig:overall}, {\ourMethod} follows a UNet-like design with encoders and decoders built upon our hybrid layers. The pipeline first voxelizes the input point clouds, processes them through hybrid layers to extract 3D features, and finally produces per-point semantic predictions through a classification head. In the following sections, we detail the key components of the hybrid layer.

\subsection{Hybrid Layer} 


The hybrid layer combines attention and Mamba operators to capture both fine-grained and global features. As shown in Fig~\ref{fig:layer}, we first divide input voxels into several equal-sized small groups for the attention layer to extract fine-grained features. Then we divide voxels into several equal-sized large groups for following the Mamba layer to extract global features. 
Besides, we use the xCPE~\cite{wu2024point} to further improve the performance. Next, we introduce each step of the hybrid layer. 

\subsubsection{Attention Layer}

The attention layer is used to extract fine-grained features in a local region. Given the features $F \in \mathbb{R}^{N\times C}$ of input voxels for the hybrid layer and group size $L$. We first divide $F$ to into non-overlapping 3D equal-sized small groups $F_g = \{F_{i}, i=1,2,3..., \lceil \frac{N}{L} \rceil\}$ by the space-filling curves~\cite{wu2024point}. Then, we apply the attention operator to extract features from these groups. Finally, we restore these groups to the shape of input voxels for the following Mamba layer. This process can be formulated as:  
\begin{equation}
\begin{aligned}
& F_g = \mathrm{Parition}(F, L), \\
& F_{g}^{'} = F_{g} + \mathrm{MSA}(F_{g}), \\
& F^{'} = \mathrm{Restore}(F_{g}^{'}, L),  \\
\end{aligned}
\end{equation}
where the $\mathrm{Parition}$, $\mathrm{Restore}$, and $MSA$ represent the partition, restore, and multi-head attention operator, respectively.

\subsubsection{Mamba Layer}

The Mamba layer extends feature learning to capture long-range feature interactions within a larger receptive field. We partition the enhanced features $F^{'} \in \mathbb{R}^{N\times C}$ into larger groups of size $K$ for global contextual modeling. Besides, we employ a bidirectional Mamba architecture to capture dependencies in both directions, enhancing the model's ability to understand long-range relationships. Following this enhancement, the processed features are restored as the same as the operation of attention layers. This process can be formulated as:
\begin{equation}
\begin{aligned}
& F^{'} = \mathrm{Parition}(F^{'}, K), \\
& F_{g}^{''} = F^{'}_{g} + \mathrm{BiMamba}(F^{'}_{g}), \\
& F^{''} = \mathrm{Restore}(F_{g}^{''}, K),  \\
\end{aligned}
\end{equation}
where the BiMamba represents the bidirectional Mamba.

\subsubsection{Inner Layer Hybrid Strategy}

Our Inner Layer Hybrid Strategy can effectively integrate Transformer and Mamba for 3D semantic segmentation. This design is based on two observations: (1) Attention operators can effectively capture relatively spatial relationships between voxels in the local region. (2) Mamba efficiently models long-range dependencies with linear complexity, but struggles with precise feature representation when extracting 3D features. Therefore, we argue that Mamba requires high-quality local features produced by attention for superior performance.

Within each Hybrid Layer, our strategy sequences the attention and Mamba layers to maximize their complementary strengths. 
The attention layer first processes input features to extract rich local features, which are then fed into the Mamba layer for efficient global context modeling. A final Feed-Forward Network (FFN) enables thorough integration of local and global information. This process is complemented by positional encoding through the xCPE module, ensuring spatial awareness throughout feature extraction.

This designed sequence ensures that the Mamba layer receives high-quality feature representations from the attention layer, while the FFN enables the effective fusion of both local and global features. 

\subsection{Discussion of Different Strategies}

\begin{figure}[ht]
\centering
\includegraphics[width=1.0\linewidth]{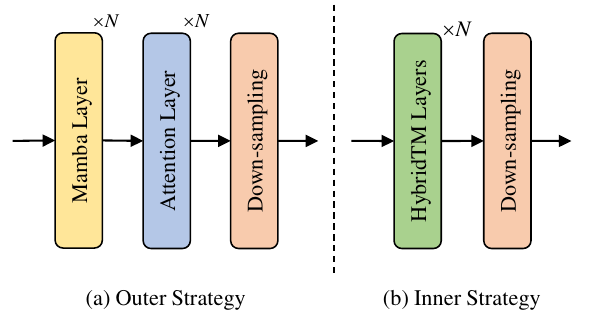}
\caption{Comparison of different hybrid strategies.}
\label{fig:strategy}
\end{figure}

The integration of attention and Mamba operators presents new challenges in 3D semantic segmentation compared to general vision tasks. In 2D vision, existing methods~\cite{hatamizadeh2024mambavision, chen2024maskmamba, liu2024map} typically adopt an outer hybrid strategy that simply allocates Mamba layers to the first $N/2$ layers and attention layers to the last $N/2$ layers within each stage (Fig.~\ref{fig:strategy}(a)). While this approach works well for dense 2D images, it proves inadequate for 3D semantic segmentation due to fundamental differences in input data characteristics.

Due to the inherently sparse and irregular nature, the point clouds must be serialized through space-filling curves~\cite{wu2024point} or window-based sorting~\cite{yang2023swin3d} for further processing. While Mamba demonstrates superior capabilities in sequence modeling for general vision tasks, it struggles to effectively capture the relative spatial relationships between voxels in these serialized sequences. This problem seriously hinders Mamba from achieving accurate segmentation performance. Unlike Transformer-based architectures that can directly model element-wise relationships through the attention operator, only adopting Mamba layers for 3D semantic segmentation leads to feature degradation. Therefore, we argue that incorporating the attention operator is essential for preserving spatial relationships and capturing fine-grained local features, while enabling Mamba to focus on modeling long-range dependencies.

As shown in Fig.~\ref{fig:strategy}, our Inner Layer Hybrid Strategy (Fig.~\ref{fig:strategy}(b)) integrates attention and Mamba layers at a fine granularity compared with Outer Strategy (Fig.~\ref{fig:strategy}(a)). By allowing attention layers to first preserve relatively spatial relationships and extract fine-grained features, we enable Mamba layers to better model global dependencies based on the fine-grained local features. Besides, we empirically demonstrate the effectiveness of the Inner Layer Hybrid Strategy in ablation studies.
\section{EXPERIMENTS}

\subsection{Datasets}

\noindent\textbf{Scannet Dataset.} ScanNet~\cite{dai2017scannet} is a large-scale dataset of 3D indoor scenes, consisting of approximately 1,500 RGB-D video sequences captured in diverse environments.  ScanNet provides extensive annotations, including 3D object instance labels and semantic segmentation for 20 categories, such as furniture and structural elements. This dataset serves as a valuable benchmark for 3D semantic segmentation in indoor settings.

\noindent\textbf{ScanNet200.} Building upon the original ScanNet framework, ScanNet200~\cite{dai2017scannet} extends the semantic annotation granularity to 200 distinct object categories. This dataset maintains the same scene split as ScanNet but presents a more challenging semantic segmentation task due to its fine-grained object categorization.

\noindent\textbf{nuScenes.} The nuScenes dataset~\cite{caesar2020nuscenes} consists of 1,000 urban driving sequences captured in Boston and Singapore. Each sequence is approximately 20 seconds long, recorded using a sensor suite of 6 cameras, 1 LiDAR, 5 radar units, and GPS/IMU. The dataset provides point-wise semantic segmentation labels for 32 classes relevant to autonomous driving scenarios, totaling 1.1B manually annotated lidar points. In our experiments, we focus on 16 major classes following previous works~\cite{wu2024point}.

\noindent\textbf{S3DIS Dataset.} The Stanford Large-Scale 3D Indoor Spaces Dataset (S3DIS)~\cite{armeni20163d} comprises detailed 3D scans of six large indoor areas across three different buildings. Collected using a terrestrial laser scanner, the dataset contains over 215 million points, each annotated with one of 13 semantic classes.

\subsection{Implementation Details}

We implement {\ourMethod} using an UNet-style architecture with asymmetric encoder-decoder paths. The encoder comprises five stages with [2,2,2,6,2] layers, while the decoder consists of four stages with [2,2,2,2] layers. For the attention mechanism, we maintain consistent group sizes of 1024 across all encoder and decoder stages. Similarly, the Mamba blocks~\cite{gu2023mamba} utilize uniform group sizes of 4096 throughout the network.



For training, we adopt the loss function and data augmentation strategies used in Point Transformer V3~\cite{wu2024point}. We optimize the network using AdamW~\cite{Loshchilov2017DecoupledWD} with the One-Cycle learning rate policy. Training is conducted on 4 NVIDIA RTX 3090 GPUs with a batch size of 12. The number of training epochs varies by dataset: 800 for both ScanNet variants, 3000 for S3DIS, and 50 for nuScenes.

\subsection{Main Results}

\begin{table}[htbp]
\caption{Comparison with state-of-the-art methods on the ScanNet validation set.}
\label{tab:scannet}
\small
\centering
\setlength{\tabcolsep}{15pt}
\resizebox{1.0\linewidth}{!}{
\begin{tabular}{l|c|c}
\toprule
Method & Present at & mIoU \\
\midrule
PointNet++~\cite{qi2017pointnet++}  & NeurIPS 2017 & 53.5 \\
MinkowskiNet~\cite{choy20194d} & CVPR 2019 & 72.2 \\
O-CNN~\cite{wang2017cnn}  &  SIGGRAPH 2017 & 74.0 \\
ST~\cite{lai2022stratified} & CVPR 2022 & 74.3 \\
Point Transformer v2~\cite{wu2022point} & NeurIPS 2022 & 75.4 \\
OctFormer~\cite{wang2023octformer}  & SIGGRAPH 2023 & 74.5 \\
Swin3D~\cite{yang2023swin3d}  & arXiv 2023 & 75.5 \\
Point Transformer v3~\cite{wu2024point} & CVPR 2024 & 77.5 \\
Pont Mamba~\cite{liu2024point} & arXiv 2024 & 75.7 \\ 
Serialized Point Mamba~\cite{wang2024serialized} & arXiv 2024 & 76.8 \\
\midrule
Ours    & -- & \textbf{77.8} \\
\bottomrule
\end{tabular}
}
\end{table}

\noindent\textbf{Results on Scannet Dataset.} In Table~\ref{tab:scannet}, we evaluate the performance of {\ourMethod} on the ScanNet validation set and compare it to several state-of-the-art (SOTA) methods. {\ourMethod} achieves 77.8\% mIoU, leading to a new SOTA result. Besides, {\ourMethod} outperforms Point Transformer v3~\cite{wu2024point} by 0.3\% mIoU. Furthermore, {\ourMethod} also outperforms other leading methods, such as Swin3D~\cite{yang2023swin3d} and Serialized Point Mamba~\cite{wang2024serialized} by 2.3\% and 1.0\% mIoU, respectively. These results demonstrate the effectiveness of {\ourMethod} to push the boundaries of 3D semantic segmentation.

\begin{table}[t!]
\caption{Comparison of our method with state-of-the-art methods on the ScanNet200 validation set.}
\begin{center}
\small
\setlength{\tabcolsep}{15pt}
\resizebox{1.0\linewidth}{!}{
\begin{tabular}{l|c|c}
\toprule
Model & Present at & mIoU \\
\midrule
MinkowskiNet~\cite{choy20194d}                 & CVPR 2019 & 25.0 \\
OctFormer~\cite{wang2023octformer}                    & SIGGRAPH 2023 & 32.6 \\
Point Transformer v2~\cite{wu2022point}         & NeurIPS 2022 & 30.2 \\
Point Transformer v3~\cite{wu2024point}         & CVPR 2024 & 35.2 \\
\midrule
Ours                         & -- & \textbf{36.5} \\
\bottomrule
\end{tabular}
}
\end{center}
\label{tab:scannet200}
\end{table}

\begin{table}[t!]
\caption{Comparison of our method with state-of-the-art methods on the nuScenes validation set}
\vspace{-15pt}
\begin{center}
\small
\setlength{\tabcolsep}{15pt}
\resizebox{1.0\linewidth}{!}{
\begin{tabular}{l|c|c}
\toprule
Model              & Present at & mIoU \\
\midrule
MinkowskiNet~\cite{choy20194d}               & CVPR 2019 & 73.3 \\
SPVNAS~\cite{tang2020searching}                       & ECCV 2020 & 77.4 \\
Cylender3D~\cite{zhu2021cylindrical}                   & CVPR 2021 & 76.1 \\
AF2S3Net~\cite{cheng20212}                     & CVPR 2021 & 62.2 \\
SphereFormer~\cite{lai2023spherical}                 & CVPR 2023 & 79.5 \\
Point Transformer v3~\cite{wu2024point}         & CVPR 2024 & 80.2 \\
\midrule
Ours                         & -- & \textbf{80.9} \\
\bottomrule
\end{tabular}
}
\end{center}
\label{tab:nusc}
\end{table}

\begin{table}[t!]
\caption{Comparison of our method with state-of-the-art methods on the S3DIS validation set.}
\vspace{-15pt}
\begin{center}
\small
\setlength{\tabcolsep}{10pt}
\resizebox{1.0\linewidth}{!}{
\begin{tabular}{l|c|c}
\toprule
Model              & Present at & Aera5 (mIoU) \\
\midrule
MinkowskiNet~\cite{choy20194d}                & CVPR 2019 & 65.4 \\
PointNeXt~\cite{qian2022pointnext}                    & NeurIPS 2022 & 70.5 \\
Swin3D~\cite{yang2023swin3d}                      & arXiv 2023 & \textbf{72.5} \\
Point Transformer v2~\cite{wu2022point}         & NeurIPS 2022 & 71.6 \\
Serialized Point Mamba~\cite{wu2024point}       & arXiv 2024 & 70.6 \\
\midrule
Ours                         & -- & 72.1 \\
\bottomrule
\end{tabular}
}
\end{center}
\label{tab:s3dis}
\end{table}

\noindent\textbf{Results on Scannet200 Dataset.} To further validate {\ourMethod} with more categories. We present a comparison of {\ourMethod} with several SOTA methods on the ScanNet200 validation set. As shown in Table~\ref{tab:scannet200}, {\ourMethod} significantly outperforms previous methods and achieves a new SOTA result (36.5\% mIoU). Specifically, 
{\ourMethod} outperforms Point Transformer V3 and OctFormer by 1.3\% mIoU and 3.9\% mIoU, respectively. These results demonstrate the effectiveness of {\ourMethod} for 3D semantic segmentation on the more complex dataset.

\noindent\textbf{Results on nuScenes Dataset.} To further validate {\ourMethod} on the outdoor dataset. In Table~\ref{tab:nusc}, we present a comparison of {\ourMethod} with several state-of-the-art methods on the nuScenes validation set. {\ourMethod} achieves 80.9\% mIoU, leading to a new SOTA result. Besides, {\ourMethod} outperforms Point Transformer v3 and SphereFormer by 0.7\% and 1.4\% mIoU, respectively. These results demonstrate the effectiveness of {\ourMethod} for handling the 3D semantic segmentation task on the outdoor large-scale dataset.

\noindent\textbf{Results on S3DIS Dataset.} We evaluate the generalization of {\ourMethod} on the smaller dataset. As shown in Table~\ref{tab:s3dis}, {\ourMethod} achieves satisfactory performance (72.1\% mIoU), which outperforms Serialized Point Mamba~\cite{wang2024serialized} by 1.5\% mIoU. These results demonstrate the generalization of {\ourMethod}.

\subsection{Ablation Studies}
We conduct ablation studies on the ScanNet validation set. The training settings are the same as the Table~\ref{tab:scannet}. 

\begin{table}[t]
\caption{Effectiveness of each layer in hybrid layer.}
\vspace{-15pt}
\setlength{\tabcolsep}{12pt}
\label{tab:effectiveness}
\begin{center}
\resizebox{1.0\linewidth}{!}{
\begin{tabular}{c|cc|c}
\toprule
\#  & Attention Layer & Mamba Layer & mIoU \\ 
\midrule
\uppercase\expandafter{\romannumeral1} & \checkmark &     & 77.1                \\ 
\uppercase\expandafter{\romannumeral2} & &  \checkmark      & 76.9                \\ 
\uppercase\expandafter{\romannumeral3} & \checkmark & \checkmark  & \textbf{77.8}       \\ 
\bottomrule
\end{tabular}
}
\end{center}
\end{table}

\begin{table}[t]
\caption{Comparison of different hybrid strategies.}
\vspace{-15pt}
\label{tab:position}
\begin{center}
\setlength{\tabcolsep}{12pt}
\resizebox{1.0\linewidth}{!}{
\begin{tabular}{c|c|c}
\toprule
\# & Strategy           & mIoU \\ 
\midrule
\uppercase\expandafter{\romannumeral1} & Outer Strategy~\cite{hatamizadeh2024mambavision} (Mamba before Attention) & 77.1 \\
\uppercase\expandafter{\romannumeral2} & Outer Strategy~\cite{hatamizadeh2024mambavision} (Mamba after Attention) & 77.4 \\
\uppercase\expandafter{\romannumeral3} & IL (Mamba before Attention)     & 77.5                \\ 
\uppercase\expandafter{\romannumeral4} & IL (Mamba after Attention)    & \textbf{77.8}       \\ 
\bottomrule
\end{tabular}
}
\vspace{-15pt}
\end{center}
\end{table}

\begin{figure*}[t]
\centering
\includegraphics[width=0.7\linewidth]{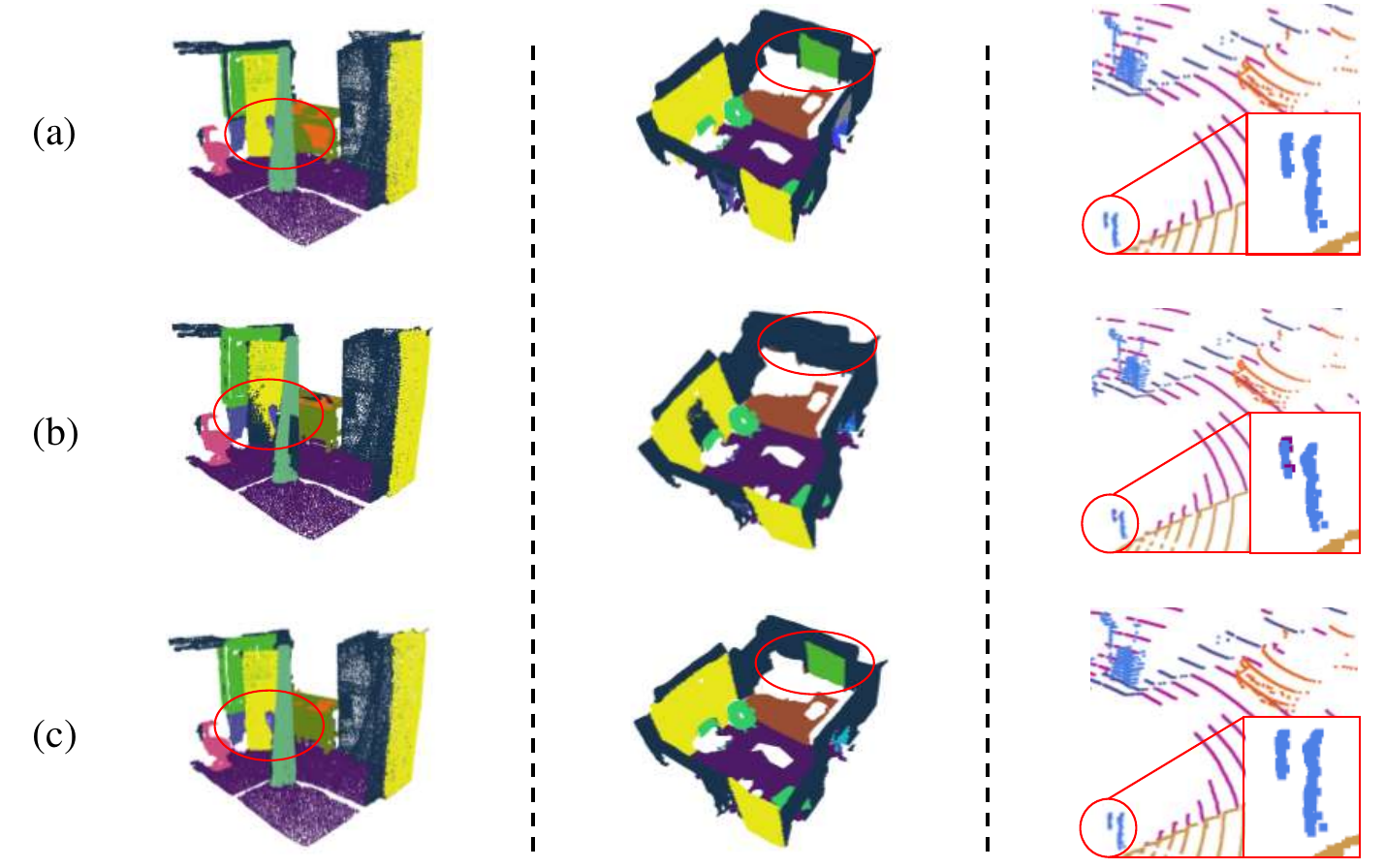}
\caption{Comparison of Point Transformer V3 (b) and {\ourMethod} (c) on the ScanNet and nuScenes validation set. (a) is the ground truth. The first and second columns are the results on ScanNet. The last column is the results on nuScenes. It can be seen that {\ourMethod} can achieve better results compared to Point Transformer V3, demonstrating the superiority of {\ourMethod}.}
\label{fig:visualization}
\end{figure*}

\noindent\textbf{Effectiveness of Components in Hybrid Layer.} We conduct ablation studies to analyze the contribution of each component within our hybrid layer architecture, with results shown in Table~\ref{tab:effectiveness}. Using only the attention layer (Configuration \uppercase\expandafter{\romannumeral1}), {\ourMethod} achieves 77.1\% mIoU. When utilizing solely the Mamba layer (Configuration \uppercase\expandafter{\romannumeral2}), performance decreases to 76.9\% mIoU, indicating that Mamba alone suffers from feature degradation. However, our proposed integration of both layers through the IL strategy (Configuration \uppercase\expandafter{\romannumeral3}) achieves 77.8\% mIoU, surpassing the single-layer variants by significant margins (0.7\% over attention-only and 0.9\% over Mamba-only). These results empirically validate the synergistic benefits of our hybrid architecture and the effectiveness of IL in combining the complementary strengths of attention and Mamba operators.




\noindent\textbf{Comparison of Different Hybrid Strategies.} To validate the effectiveness of our Inner Layer Hybrid Strategy (IL), we compare different hybrid strategies. As shown in Table~\ref{tab:position}, the Outer Strategy (Mamba before Attention) used in MambaVision~\cite{hatamizadeh2024mambavision} (\uppercase\expandafter{\romannumeral1}) can only achieve 77.1\% mIoU. When we place the attention layer before Mamba in Outer Strategy (\uppercase\expandafter{\romannumeral2}), {\ourMethod} can achieve 77.4\%, which outperforms (\uppercase\expandafter{\romannumeral1}) by 0.3\% mIoU. These results demonstrate that attention should be placed before Mamba. When we use IL (Mamba before Attention) (\uppercase\expandafter{\romannumeral3}), {\ourMethod} can achieve 77.5\% mIoU, which outperforms (\uppercase\expandafter{\romannumeral1}) by 0.4\% mIoU. When we use IL (Mamba after Attention) (\uppercase\expandafter{\romannumeral4}), {\ourMethod} can achieve 77.8\% mIoU, which outperforms (\uppercase\expandafter{\romannumeral1}), \uppercase\expandafter{\romannumeral2}, (\uppercase\expandafter{\romannumeral3}) by 0.7\%, 0.4\%, 0.5\% mIoU, respectively. These results demonstrate the superiority of our IL compared to other strategies.




\subsection{Visualization}

To illustrate the superiority of {\ourMethod}, we present the visualization of the qualitative results of Point Transformer V3~\cite{wu2024point} (b) and {\ourMethod} (c) on the ScanNet and nuScenes validation set in Fig.~\ref{fig:visualization}. In the first and second columns, {\ourMethod} can achieve more precise segmentation for large objects. In the last column, benefiting from the proposed IL hybrid strategy, {\ourMethod} can also accurately segment the small object. The visualization results demonstrate that our method not only focuses on local details but also attends to broader regions, leading to better overall segmentation performance.


\section{CONCLUSION}
In this paper, we present {\ourMethod}, the first hybrid architecture for 3D semantic segmentation that synergistically combines Transformer and Mamba. In addition, we propose the Inner Layer Hybrid Strategy, enabling effective integration of fine-grained spatial feature extraction and efficient global context modeling, specifically designed for 3D semantic segmentation. The effectiveness of our approach is validated through state-of-the-art performance on challenging indoor (ScanNet, ScanNet200) and outdoor (nuScenes) benchmarks, demonstrating the superiority of our hybrid architecture for 3D semantic segmentation.

\bibliographystyle{IEEEtran}
\bibliography{references}

\end{document}